\pdfoutput=1

\documentclass[11pt]{article}

\usepackage[final]{coling}
\usepackage{todonotes}
\usepackage{times}
\usepackage{latexsym}
\usepackage{makecell}
\usepackage[most]{tcolorbox}
\usepackage{xcolor}

\definecolor{MutedGreen}{rgb}{0.45, 0.55, 0.25}  
\definecolor{MutedRed}{rgb}{0.7, 0.1, 0.1}       

\definecolor{darkraspberry}{rgb}{0.53, 0.15, 0.34}
\definecolor{olive}{rgb}{0.5, 0.5, 0.0}

\usepackage[T1]{fontenc}

\usepackage[utf8]{inputenc}

\usepackage{microtype}
\usepackage{rotating} 
\usepackage{adjustbox} 
\usepackage{booktabs} 
\usepackage{inconsolata}

\usepackage{graphicx}

\usepackage{xspace}
\usepackage{booktabs}
\usepackage{multicol}
\usepackage{xcolor}
\usepackage{float}

\newcommand{\anli}{\textsc{\textbf{anli}}\xspace}
\newcommand{\snli}{\textsc{\textbf{snli}}\xspace}
\newcommand{\xnli}{\textsc{\textbf{xnli}}\xspace}
\newcommand{\pawsx}{\textsc{\textbf{pawsx}}\xspace}
\newcommand{\stshard}{\textsc{\textbf{sts-h}}\xspace}
\newcommand{\stseasy}{\textsc{\textbf{sts}}\xspace}
\newcommand{\sickeasy}{\textsc{\textbf{sick}}\xspace}

\newcommand{\amr}{\textsc{\textbf{true}}\xspace}
\newcommand{\simp}{\textbf{\textsc{simp}}\xspace}
\newcommand{\mrpc}{\textbf{\textsc{mrpc}}\xspace}

\newcommand{\ourbenchmark}{\textbf{\textsc{para\-phra\-sus}}\xspace}

\definecolor{antiquefuchsia}{rgb}{0.57, 0.36, 0.51} 
\definecolor{azure(colorwheel)}{rgb}{0.0, 0.5, 1.0}
\newcommand{\pa}{\textcolor{azure(colorwheel)}{P1}} 
\newcommand{\pb}{\textcolor{azure(colorwheel)}{P2}} 
\newcommand{\pc}{\textcolor{azure(colorwheel)}{P3}} 
\newcommand{\paicl}{\textcolor{azure(colorwheel)}{P1-ICL\_K4}} 

\newcommand{\pbicl}{\textcolor{azure(colorwheel)}{P2-ICL\_K4}} 

\newcommand{\pcicl}{\textcolor{azure(colorwheel)}{P3-ICL\_K4}} 

\newcommand{\icl}{\textcolor{azure(colorwheel)}{\-ICL\_K4}} 

\title{PARAPHRASUS: A Comprehensive Benchmark for Evaluating Paraphrase Detection Models}

\author{Andrianos Michail \\
    University of Zurich \\
    \texttt{andrianos.michail@cl.uzh.ch} \And
    Simon Clematide \\
    University of Zurich \\
    \texttt{simon.clematide@cl.uzh.ch} \\ \AND 
    Juri Opitz \\
    University of Zurich \\
    \texttt{jurialexander.opitz@cl.uzh.ch}
}

\begin{document}
\maketitle
\begin{abstract}
The task of determining whether two texts are paraphrases has long been a challenge in NLP. However, the prevailing notion of paraphrase is often quite simplistic, offering only a limited view of the vast spectrum of paraphrase phenomena. Indeed, we find that evaluating models in a paraphrase dataset can leave uncertainty about their true semantic understanding. To alleviate this, we create \ourbenchmark,
a benchmark designed for multi-dimensional assessment, benchmarking and selection of paraphrase detection models. We find that paraphrase detection models under our fine-grained evaluation lens exhibit trade-offs that cannot be captured through a single classification dataset. Furthermore, \ourbenchmark allows prompt calibration for different use cases, tailoring LLM models to specific strictness levels. \ourbenchmark includes 3 challenges spanning over 10 datasets, including 8 repurposed and 2 newly annotated; we release it along with a benchmarking library at \url{https://github.com/impresso/paraphrasus}

\end{abstract}

\section{Introduction}

Our study was set in motion by a serendipitous finding. Like many other researchers at the time, we were benchmarking  large language models (LLMs). We had \textit{particular interest in the paraphrasing detection task}\footnote{We consider paraphrasing as an important and challenging task, hand in hand with \citet[p. 117]{quintilus} who lived 2,000 years ago, and of course, with many papers presented at CL and ML conferences such as \citet{bhagat2013paraphrase, zhou-bhat-2021-paraphrase, NEURIPS2023_575c4500}.} and evaluated Llama and others LLMs \citep{dubey2024llama, alves2024tower} on the adversarial paraphrase detection test set \textit{PAWS-X} \citep{yang-etal-2019-paws}. Interestingly, LLMs appear to under-perform much on \textit{PAWS-X}. Their performance lies slightly above random coin flip, and they are vastly outperformed by smaller BERT-based models fine-tuned on the training split of \textit{PAWS-EN}.
\begin{figure}[ht]
    \centering    \includegraphics[width=0.49\textwidth]{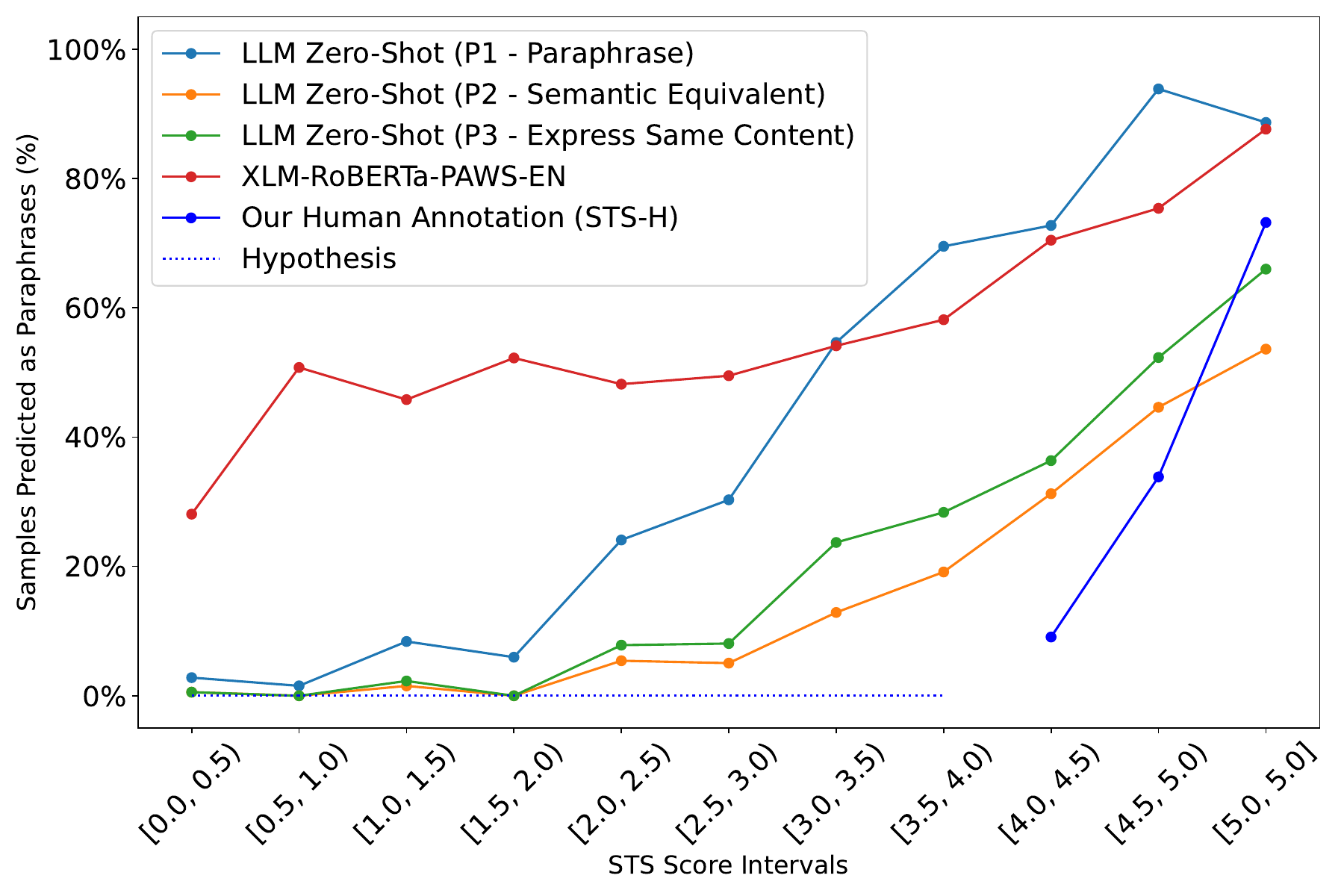} 
    \caption{Percentage of paraphrases predicted on the Semantic Text Similarity Dataset (STSBenchmark) dataset \citep{cer-etal-2017-semeval}, binned by scores from 0 (completely dissimilar) to 5 (completely equivalent). Human annotation comes from the \stshard human annotation we perform.}
    \label{fig:stsbenchmark_curve}
\end{figure}

Therefore, we investigated whether from analyzing results on PAWS-X, we could make conclusions about the paraphrase detection understanding (or lack thereof) of different models. As what was merely intended as a sanity check at first, we repurposed data from the semantic text similarity (STS) task, and investigated the distribution of predicted paraphrases against the fine-grained similarity labels, where only the highest label of 5 denotes true paraphrases. In fact, moving away from 5, the semantic similarity decreases rapidly, so there clearly are no paraphrases below a certain level (e.g., 4). 
We confronted different models with this simple test and were surprised by the results (see Figure~\ref{fig:stsbenchmark_curve}). Suddenly, the predictions of LLMs appeared more reasonable than what we took away from the benchmarking results, whereas the trained models that performed well on PAWS-X appeared na\"ive and much too over-confident, assigning many paraphrase predictions to sentences that are semantically  different. Indeed, ``predicting paraphrases is not easy'' \citep{vahtola-etal-2022-easy}. 

Inspired by these pilot findings, we began a deeper empirical exploration into paraphrase notions and paraphrase detection models, the results of which are presented in the remainder of this paper. A main outcome of this is our \ourbenchmark benchmark that allows the community to test paraphrase detection models in different aspects, building on existing, novel, and carefully repurposed data, eliciting diverse notions of paraphrases. With  \ourbenchmark, we can gather a fine-grained picture of paraphrase modes performance that reveals particular strengths and weaknesses. Our contributions are the following:





\begin{enumerate}
\item We present \ourbenchmark, a multi-faceted evaluation benchmark for paraphrase detection, including  datasets of human-written sentence pairs of varying semantic and lexical similarity repurposed for testing the generalization of paraphrase detection models.
\item Within our benchmark, we also contribute two novel datasets. i) We annotate a test set of 338 semantically similar sentence pairs for paraphrase classification, creating a challenging test set that paraphrase detection methods plateau at 60\% accuracy. ii) We exploit guidelines for Abstract Meaning Representation annotations, extracting expert examples of true paraphrases.
\item To showcase the usefulness of \ourbenchmark, we test LLMs and trained models under different setups, revealing new insights about paraphrase detection models and paraphrase data. A main insight is that evaluating a model on a single paraphrase dataset, as commonly done before, can yield a misleading picture about actual generalization performance. The effect is especially marked for trained models.
\item We perform quantitative and qualitative analyses, learning about suitable model training regimes and human perception of paraphrases.
\end{enumerate}

\section{Related Work}
\label{sec:rw}

\paragraph{Paraphrasing.} We find early excitement about paraphrases in the rhetorician \citet{quintilus}'s book, written  2000 years ago. \citeauthor{quintilus} suggests paraphrasing as an  exercise for classrooms, praising the task as ``valuable in virtue of its difficulty'', the goal is ``to rival and vie with the original in the expression of the same thoughts'' (p. 117). A useful taxonomy of different types of paraphrases can be found later in \citet{bhagat2013paraphrase}'s work, outlining 25 means of creating one, e.g., by metaphor substitution or function word variation (``Pat gave a nice demo'', ``Pat's demo was nice''). 

Clearly, there are many NLP applications where paraphrasing and understanding paraphrases plays a crucial role \citep{mallinson-etal-2017-paraphrasing}. Some of these are semantic search \citep{reimers-gurevych-2019-sentence}, style transfer \citep{krishna-etal-2020-reformulating}, machine translation \citep{madnani-etal-2007-using}, plagiarism \citep{barron-cedeno-etal-2013-plagiarism, sharjeel-etal-2016-uppc}, and text reuse \citep{buchler2012increasing}. The evaluation of text generation is often interested in whether a candidate and the reference are paraphrases \citep[e.g.,][]{freitag-etal-2020-human, thompson-post-2020-automatic, nawrath-etal-2024-role}.

\paragraph{Paraphrase Datasets}

Recognizing the importance of paraphrasing, prior research has developed methods for constructing datasets, including mining large corpora for multilingual paraphrases \citep{ganitkevitch-etal-2013-ppdb, ganitkevitch-callison-burch-2014-multilingual} and using back-translation to generate 50 million paraphrase pairs \citep{wieting-gimpel-2018-paranmt}. Such large-scale datasets are especially attractive for training models for each of the diverse paraphrase-related applications mentioned above. On the other hand, smaller datasets have been proposed, with a stronger focus on quality and evaluation of paraphrase judgment models. One of the earliest of such datasets is the \mrpc corpus \citep{dolan-brockett-2005-automatically}, this paraphrase classification dataset was constructed by mining the web and filtering the most promising candidates using lexicon-based classifiers. Two non-specialist human judges then labelled the sentence pairs, resulting in a test set of 1,730 sentence pairs, of which 66.5\% were labelled positive.

Recently, the PAWS dataset \citep{zhang-etal-2019-paws} was introduced, providing an adversarial dataset for paraphrase classification through the application of word scrambling.  They create challenging pairs by controlled word swapping and back-translation, followed by human quality check for fluency and paraphrase semantic. The dataset is split in 49,401 training sentence pairs and 8,000 test sentence pairs. \pawsx \citep{yang-etal-2019-paws} takes 2000 of the PAWS Wiki test samples and creates a 7-way multilingual test set in German, Spanish, French, Chinese, Japanese, and Korean using human translators. While these minimal pairs are interesting to perform a focused evaluation, its adversarial mode of dataset creation can introduce limitations on the coverage of linguistic formulation variation and hinder generalization abilities. 

Typically, there is a quantity-quality tradeoff. With \ourbenchmark we aim to get the best of the worlds of quantity and quality, releasing a comprehensive and trustworthy multifold dataset for evaluating paraphrase detection models and analyzing various sorts of paraphrases.

\paragraph{Creating benchmarks.} Benchmarking is a very active topic in NLP \cite{gardent-etal-2017-webnlg, zhu2018texygen}, and a key tool for study and development of LLMs \citep{srivastava2023beyond}. These benchmarks are useful since they help to assess and select models within a broader picture, and thus encourage their further development \cite{gehrmann-etal-2021-gem}. A recent critique that has emerged is due to the way how LLMs are trained, as they are often trained on large corpora crawled from the web. Hence, benchmark data may have leaked into the model, and the performance is harder to assess \citep{deng2024benchmark}. A way to at least partially mitigate this concern is by repurposing data from related tasks. Specifically, while \citet{HonovichAharoni:2022} repurpose data for their factuality benchmark from existing factuality datasets from different domains, we take a step further and source data not only from different domains, but also from related tasks, projecting them as paraphrase classification. This further increases trust in evaluation results based on our proposed \ourbenchmark benchmark.

\paragraph{Semantic similarity datasets.} STS \citep{agirre-etal-2013-sem,cer-etal-2017-semeval} and SICK \cite{marelli-etal-2014-sick} elicited human ratings of sentence similarity on a Likert scale. While STS annotates \textit{semantic} similarity, SICK annotates \textit{semantic} relatedness. These two aspects are highly related, but not exactly the same \cite{budanitsky2006evaluating, kolb-2009-experiments}. However, the highest scores on the Likert scales of SICK and STS denote the equivalence of meaning of two sentences. We exploit this property for our \ourbenchmark paraphrase benchmark by carefully selecting  lower similarity pairs as non-paraphrases for controlled tests.

\section{Proposed Benchmark}

\begin{table}[t]
\centering
\resizebox{\columnwidth}{!}{ 
\begin{tabular}{l r r r r r}
\toprule
\textbf{Dataset} & \textbf{Pairs} & \multicolumn{2}{c}{\textbf{Paraphrase}} & \multicolumn{2}{c}{\textbf{$\neg$Paraphrase}} \\
\cmidrule(lr){3-4} \cmidrule(lr){5-6}
 &  & \textbf{abs.} & \textbf{rel.} & \textbf{abs.} & \textbf{rel.} \\
\midrule
\multicolumn{6}{c}{\textbf{Classify!}} \\
\pawsx     & 14,000 & 6,160  & 44\%  & 7,840  & 56\% \\
\mrpc        & 1,730  & 1,159  & 67\%  & 571    & 33\% \\
\stshard    & 338    & 109    & 32\%  & 229    & 68\% \\
\midrule
\textbf{Group} & \textbf{16,068} & \textbf{7,427} & \textbf{48\%} & \textbf{8,641} & \textbf{52\%} \\
\midrule
\multicolumn{6}{c}{\textbf{Minimize!}} \\
\snli       & 6,632  & 0      & 0\%   & 6,632  & 100\% \\
\anli        & 798    & 0      & 0\%   & 798    & 100\% \\
\xnli        & 16,700 & 0      & 0\%   & 16,700 & 100\% \\
\stseasy    & 706    & 0      & 0\%   & 706    & 100\% \\
\sickeasy   & 2,305  & 0      & 0\%   & 2,305  & 100\% \\\midrule
\textbf{Group} & \textbf{27,141} & \textbf{0} & \textbf{0\%} & \textbf{27,141} & \textbf{100\%} \\
\midrule
\multicolumn{6}{c}{\textbf{Maximize!}} \\
\amr         & 167    & 167    & 100\% & 0      & 0\% \\
\simp   & 600    & 600    & 100\% & 0      & 0\% \\\midrule
\textbf{Group} & \textbf{767} & \textbf{767} & \textbf{100\%} & \textbf{0} & \textbf{0\%} \\
\midrule
\textbf{Total} & \textbf{43,976} & \textbf{8,194} & \textbf{34\%} & \textbf{35,782} & \textbf{66\%} \\
\bottomrule
\end{tabular}
}
\caption{Overview of the datasets in the \ourbenchmark benchmark along with their label distribution.}
\label{tab:benchmark_overview}
\end{table}
Paraphrases  come in different flavors, and distinguishing types of paraphrases can be  fuzzy \citep{bhagat2013paraphrase}. Our benchmark therefore adopts an empirical approach to evaluating paraphrase detection models, according to three desiderata: i) It should reflect a broad spectrum of domains, to lower the impact of any superficial training that may have been acquired by a model. ii) It should reflect a broad spectrum of different flavors of paraphrases, e.g., varying strictness levels of paraphrases, negative examples with contradiction or neutral semantic relation. iii) It should help us select models for different paraphrase challenges, or select a single model that performs consistently across multiple categories. Table~\ref{tab:benchmark_overview} gives an overview of the size and label distribution of each of the ten evaluation sets. More statistics are available in Appendix~\ref{sec:descriptive}.

\subsection{The data: Ten Parts with Three Objectives}

Our benchmark consists of 10 parts, with eight of them repurposed from various types of NLP tasks, such as natural language inference (NLI) and meaning representation annotation guidelines. By distributing the data across different domains, we aim to capture a wide range of paraphrase phenomena. An additional benefit of repurposing existing data could be that it helps alleviate concerns that the LLMs being evaluated may have incorporated this data during training. In addition to the repurposed data, three datasets are genuine paraphrase datasets, one of which is a human-annotated set of particularly challenging examples created by us.

The benchmark is divided into three challenges, along which we structure its description:

\paragraph{Classify!} We evaluate the models on three binary paraphrase classification datasets, where each model must determine whether a given pair of sentences constitutes a paraphrase or not.

\begin{itemize}
    \item \pawsx: As mentioned in Section \ref{sec:rw}, \pawsx is a multilingual dataset containing lexically similar pairs of paraphrases and non-paraphrases. 
    \item \mrpc: As also mentioned in Section \ref{sec:rw}, \mrpc is one of the first paraphrase detection dataset that has been released \citep{dolan-brockett-2005-automatically}.
    
    \item \stshard: We propose using challenging sentence pairs without relying on adversarial strategies. To achieve this, we select highly similar sentence pairs from the STS dataset, which are annotated on a 5-point Likert scale. Potential paraphrase candidates are situated at the upper end of this scale. We select all pairs from 4-5 and carefully create a high-quality set of 338 paraphrase annotations.%
 
    A semantics expert and a student annotated this subset independently, resulting in moderate-to-good Kappa of 0.63 \cite{Cohen:1960}. Finally, disagreements were discussed and adjudicated. For the few cases where  disagreements could not be resolved (13  out of 56  total), we assigned the non-paraphrase class. This decision was based on the reasoning that it is easier to argue that two sentences are not paraphrases than to prove the opposite.
\end{itemize} 

\paragraph{Minimize positive predictions!} 
In a dataset that contains no paraphrases, the primary objective of the model is to minimize the prediction of paraphrases. However, to ensure the task remains challenging, the paired texts should exhibit some degree of similarity. In this study, we repurpose the following five datasets.

\begin{itemize}
    \item  \snli, \anli, \xnli --- Natural Language Inference (NLI) repurposed: NLI asks whether a hypothesis follows a given premise, labeling the relationship with \textit{entailed, neutral, or contradiction}. We select those pairs that either stand in a neutral relationship, or in a contradiction. Since typically the premise is longer than the hypothesis, we control this bias by adding the flipped pairs. The resulting data are denoted by \snli, sampled from the first large-scale NLI dataset \citep{bowman-etal-2015-large}; \anli taken from an adversarial version of the task \citep{nie-etal-2020-adversarial}; and the cross-lingual \xnli\cite{conneau-etal-2018-xnli} \footnote{Limited to EN, DE, FR, ES, ZH due to compute costs.} that allows us to study paraphrase detection models in the minimization task in different languages.
    \item \stseasy, \sickeasy --- Similarity repurposed: Orthogonally to how we created the \stshard data by using only extremely similar pairs, we now select only pairs that we know that they are not paraphrases. To ensure maximum data quality and lower the possibility of annotation confusions in the upper spectrum of the similarity Likert scale, we select only values 0-3 as negative pairs, excluding the range of 3 to 5. The resulting data are gathered from two datasets \citep{marelli-etal-2014-sick, cer-etal-2017-semeval}.
\end{itemize}

\paragraph{Maximize positive predictions!} If we know that a dataset contains only pairs of paraphrases, the natural expectation towards a model would be to maximize its paraphrase detection rate. We selected two datasets to test this behavior:

\begin{itemize}
\item \amr: We create a dataset of simple, guaranteed paraphrases. The objective for a paraphrase model using this dataset is to correctly identify as many paraphrases as possible.  To ensure the accuracy of these examples, we  leverage Abstract Meaning Representation annotation guidelines \citep{banarescu-etal-2013-abstract}.\footnote{\url{https://github.com/amrisi/amr-guidelines}} AMR has the goal of mapping the same meaning to the same structure, and highlights this in its annotation guidelines by presenting graphs together with a group of semantically equivalent (but structurally different) sentences. For every meaning graph, we extract the $n$-sized set of example sentences and create $\frac{n^2 -n}{2}$ true paraphrase pairs.\footnote{Since AMR currently captures many, but not all semantic aspects \citep{sadeddine-etal-2024-survey}, we exclude any parts of the annotation guidelines where AMR conflates two aspectually divergent sentences into a single representation.} As the guidelines succinctly treat diverse phenomena, there can be short phrases, like ``20 km'' and ``20 kilometers'', but most of them are short sentences, e.g., ``The boy desires the girl to believe him.'' and ``The boy has a desire to be believed by the girl.'' \footnote{The license of the AMR guidelines allows us to publicly release this  subset of true paraphrases.} 

\item \simp: From the AMR guidelines, we retrieved simple and accurate paraphrases; however, the manually constructed nature of the data limits its diversity. To create a more diverse dataset of paraphrases, albeit at the cost of introducing more noise, we repurpose parallel sentence pairs of different readability levels (Advanced, Intermediate, Elementary) from \citet{vajjala-lucic-2018-onestopenglish}'s  OneStopEnglish.
\end{itemize}

\begin{table*}[t]
\centering
\small
\resizebox{\textwidth}{!}{ 
\begin{tabular}{l|r|r|r|r|r|r|r|r|r|r||r|r|r|r}
\toprule
Results are error percentages $\boldsymbol{\downarrow}$ & \multicolumn{3}{c|}{\textbf{Classify!}} 
& \multicolumn{5}{c|}{\textbf{Minimize!}} 
& \multicolumn{2}{c||}{\textbf{Maximize!}} 
& \multicolumn{4}{c}{\textbf{\textit{Averages}}}  \\
\midrule
\textbf{Classification Method} 
& \rotatebox[origin=c]{90}{~~\pawsx} 
& \rotatebox[origin=c]{90}{\stshard} 
& \rotatebox[origin=c]{90}{\mrpc} 
& \rotatebox[origin=c]{90}{\snli} 
& \rotatebox[origin=c]{90}{\anli} 
& \rotatebox[origin=c]{90}{\xnli} 
& \rotatebox[origin=c]{90}{\stseasy} 
& \rotatebox[origin=c]{90}{\sickeasy}
& \rotatebox[origin=c]{90}{\amr} 
& \rotatebox[origin=c]{90}{\simp} 
& Clfy! & Min! & Max! & \textbf{$\overline{Err}$}\\
\midrule %
XLM-R$\leftarrow$ \textit{PAWS-EN Train} 
& \textbf{15.2} & 54.1 & 33.4 & 32.4 & 7.2 & 26.7 & 46.6 & 37.0 & 31.4 & \textbf{5.3} & \textbf{34.2} & 30.0 & 18.3 & 27.5 \\ 
Llama3 Instruct \pa  
& 44.7 & 56.2 & \textbf{23.6} & 7.3 & 13.0 & 12.3 & 12.9 & 0.9 & \textbf{9.0} & 14.7 & 41.5 & 9.3 & \textbf{11.8} & \textbf{$^\star$20.9} \\
Llama3 Instruct \pb 
& 40.7 & \textbf{37.6} & 45.9 & 1.0 & 1.2 & 1.4 & 2.4 & 0.1 & 34.7 & 47.3 & 41.4 & 1.2 & 41.0 & 27.9 \\
Llama3 Instruct \pc
& 38.1 & 41.7 & 37.5 & 1.3 & 1.7 & 1.3 & 3.5 & \textbf{0.0} & 35.3 & 37.5 & 39.1 & 1.6 & 36.4 & 25.7 \\
Llama3 Instruct \paicl
& 39.0 & 44.7 & 33.2 & 1.9 & 2.0 & 2.8 & 3.5 & 0.3 & 29.9 & 33.3 & 39.0 & 2.1 & 31.6 & 24.2 \\
Llama3 Instruct \pbicl
& 34.1 & 41.7 & 45.2 & 0.8 & \textbf{0.8} & \textbf{0.3} & 3.1 & \textbf{0.0} & 40.1 & 42.3 & 40.3 & 1.0 & 41.2 & 27.5 \\
Llama3 Instruct \pcicl
& 33.2 & 39.1 & 46.7 & \textbf{0.5} & \textbf{0.8} & \textbf{0.3} & \textbf{2.4} & \textbf{0.0}  & 50.9 & 45.5 & 39.7 & \textbf{0.8} & 48.2 & 29.6 \\
\bottomrule
\end{tabular}}
\caption{Main results on \ourbenchmark. For description of prompts \pa, \pb, \pc, and their ICL variants, see \S \ref{subsec:llm}. All results are error percentages: Lower numbers are better. The classification method with the best overall performance is marked with an asterisk ($^\star$).}
\label{tab:mainresults}
\end{table*}

\subsection{Evaluation Metric}

To achieve a final ranking of systems, we need to select a metric. While there are many choices for a metric, and each has their benefits and drawbacks \citep{10.1162/tacl_a_00675}, we wish for an interpretable and robust measure. For each dataset, we calculate a classification $error$ percentage. For datasets from the classify challenge, this is the ratio of wrong predictions. For datasets where the goal is to minimize the amount of positive predictions, like the neutral and contradictory pairs of the NLI data, this is the ratio of positive predictions (in that case, equivalent to false positive rate). For the datasets in the maximization challenge, such as \amr, this is the ratio of negative predictions (equivalent to false negative rate). To achieve a robust final score, we calculate an unweighted average per each of the three challenges, and then another unweighted average over these averages:
\begin{equation*}
  \overline{Err} = \frac{\sum\limits_{\text{challenge} \in \text{benchmark}} \left( \frac{\sum\limits_{\text{dataset} \in \text{challenge}} \text{error(dataset)}}{|\text{challenge}|} \right)}{|\text{benchmark}|}
\end{equation*}
The $\overline{Err}$ models the probability of observing a wrong prediction, given we randomly select one of the three challenges, and randomly pick an example from a random dataset of the task. Since all scores on \ourbenchmark are error measurements (or their averages), lower numbers are better.

\section{Testing Paraphrase Detection Models} 

\begin{figure}[t]  
\begin{tcolorbox}[title={Zero Shot Paraphrase Detection Prompt},label={prompt-zs_main},colback=white]
Are the following sentences \{paraphrase notion\}? \newline
\newline
Sentence 1: \{sentence1\} \newline
Sentence 2: \{sentence2\} \newline
\newline
Answer with 'Yes' or 'No'
\end{tcolorbox}
\caption{For \pa{}, \pb{}, and \pc{}, the paraphrase notions we ask for are ``paraphrases'', ``semantically equivalent'' and ``expressing the same content'' respectively. For the ICL expanded prompt, see Appendix~\ref{sec:prompts}.}
\label{fig:prompt_zs_main}
\end{figure}

In this section, we introduce the methods and models used for systematically testing our \ourbenchmark benchmark to assess the paraphrase detection capabilities. We focus on two research questions:

\begin{enumerate}
    \item Given that \pawsx is a widely used dataset for measuring the Natural Language Understanding of LLMs and is relatively large, we ask: What insights can be gained from training on it? Specifically, we aim to determine how useful \pawsx is for training smaller, more efficient models.
    \item Given the increasing use of LLMs, we aim to test a reference open-source LLM to examine its ability to recognize different types of paraphrasing in \ourbenchmark. We also explore different ways of describing the concept of paraphrasing to assess the best methods for prompting LLMs to classify paraphrases.
\end{enumerate}

\subsection{What can we learn from \pawsx? 
}

We replicate the approach from the original \pawsx study by \citet{yang-etal-2019-paws} and fine-tune a multilingual encoder,  XLM-RoBERTa $_{base}$ (\textbf{XLM-R}) \cite{conneau-etal-2020-unsupervised} on the PAWS Wiki English training set. Using default fine-tuning hyperparameters, we train for 6 epochs. During initial evaluation, we observed considerable variance between different fine-tuning seeds on \ourbenchmark. To reduce the impact of this variance on our results, we report the average performance based on three high-performing checkpoints, each corresponding to a different seed, yielding a total of nine predictions.

\subsection{What can we learn from LLMs?}%
\label{subsec:llm}
Recent research has indicated that LLMs have developed a strong understanding of human language, which can be accessed through natural language prompts. In this study, we prompt Llama3 Instruct 8B \citep{dubey2024llama} quantized to 4 bits with three different paraphrase notions  along with their in-context learning equivalents.


\paragraph{Expressing different paraphrase notions.} Given that multiple ``personalities'' can be elicited from LLMs \citep{chan2024chateval}, we leverage this property to design three distinct, minimalistic  prompts that vary only in how the concept of ``paraphrase'' expressed. The first prompt (henceforth denoted as \pa{}) is most straightforward, just asking the model to judge whether two sentences are \textit{paraphrases}. The second prompt \pb{} intends to emulate a person with background in semantics, asking the model whether the following two sentences are \textit{semantically equivalent}. We speculate that this prompt triggers a model mode that very strictly judges the pairs, refraining from assigning a positive label if there is only a minor semantic difference. The third prompt \pc{}  aims to kindle a person who is primed from Information Retrieval: The model is asked whether the following two sentences are about \textit{the same content}. We speculate that this prompt provides a suitable balance between the straightforward \pa{} prompt and the presumed strict prompting \pb.

\paragraph{In-context learning with PAWS-X examples.} Our second version of these prompt personas leverages the ``emergent capability'' of in-context learning \citep{brown2020language}, and is henceforth denoted by the \icl{} suffix. Compared to the original, during inference time, we sample two positive and two negative sentence pairs from the training set of PAWS-Wiki and prepend them to the prompt.

\begin{table*}[t] 
\centering
\small 
\resizebox{\textwidth}{!}{ 
\begin{tabular}{l|r|r|r|r|r|r|r|r|r|r||r|r|r|r}
\toprule
Results are error percentages $\boldsymbol{\downarrow}$ & \multicolumn{3}{c|}{\textbf{Classify!}} 
& \multicolumn{5}{c|}{\textbf{Minimize!}} 
& \multicolumn{2}{c||}{\textbf{Maximize!}} 
& \multicolumn{4}{c}{\textbf{\textit{Averages}}}  \\
\midrule
\textbf{Model} $\leftarrow$ \textit{PAWS-EN \% Negatives to EasyNegs.}
& \rotatebox[origin=c]{90}{\pawsx} 
& \rotatebox[origin=c]{90}{\stshard} 
& \rotatebox[origin=c]{90}{\mrpc} 
& \rotatebox[origin=c]{90}{\snli} 
& \rotatebox[origin=c]{90}{\anli} 
& \rotatebox[origin=c]{90}{\xnli} 
& \rotatebox[origin=c]{90}{\stseasy} 
& \rotatebox[origin=c]{90}{\sickeasy}
& \rotatebox[origin=c]{90}{\amr} 
& \rotatebox[origin=c]{90}{\simp} 
& Clfy! & Min! & Max! & \textbf{$\overline{Err}$}\\
\midrule %
\textbf{XLM-R} $\leftarrow$ \textit{Original Negatives} (\textcolor{MutedGreen}{44$\%$} | \textcolor{MutedRed}{56$\%$})    
& 15.2 & \textbf{54.1} & 33.4 & \textbf{32.4} & 7.2 & 26.7 & 46.6 & 37.0 & 31.4 & 5.3 & 34.2 & 30.0 & 18.3 & 27.5 \\
\textbf{XLM-R} $\leftarrow$ \textit{25\% Easy Negatives} (\textcolor{MutedGreen}{44$\%$} | \textcolor{MutedRed}{56$\%$}) 
& \textbf{15.1} & 54.8 & 32.1 & 32.8 & \textbf{5.9} & \textbf{20.8} & \textbf{42.6} & \textbf{32.7} & 32.0 & 4.0 & \textbf{34.0} & \textbf{27.0} & 18.0 & $^\star$\textbf{26.3} \\
\textbf{XLM-R} $\leftarrow$ \textit{50\% Easy Negatives} (\textcolor{MutedGreen}{44$\%$} | \textcolor{MutedRed}{56$\%$})
& 15.9 & 58.5 & \textbf{30.8} & 43.6 & 7.7 & 29.8 & 58.9 & 50.7 & 14.5 & 2.3 & 35.1 & 38.1 & 8.4 & 27.2 \\ 
\textbf{XLM-R} $\leftarrow$ \textit{75\% Easy Negatives} (\textcolor{MutedGreen}{44$\%$} | \textcolor{MutedRed}{56$\%$}) 
& 18.0 & 63.1 & 33.2 & 75.8 & 38.0 & 67.4 & 85.3 & 78.9 & \textbf{5.3} & \textbf{1.1} & 38.1 & 69.1 & \textbf{3.2} & 36.8 \\
\bottomrule
\end{tabular}}
\caption{Ablation study A: XLM-R fine-tuned on PAWS EN Wiki training samples where a percentage of the second sentences of the negative pairs was exchanged with a random sentence. All results are error percentages: Lower numbers are better.  The classification method with the best overall performance is marked with an asterisk ($^\star$).}
\label{tab:ablation_negatives_paws}
\end{table*}

\begin{table*}[t] 
\centering
\small 
\resizebox{\textwidth}{!}{ 
\begin{tabular}{l|r|r|r|r|r|r|r|r|r|r||r|r|r|r}
\toprule
Results are error percentages $\boldsymbol{\downarrow}$ & \multicolumn{3}{c|}{\textbf{Classify!}} 
& \multicolumn{5}{c|}{\textbf{Minimize!}} 
& \multicolumn{2}{c||}{\textbf{Maximize!}} 
& \multicolumn{4}{c}{\textbf{\textit{Averages}}}  \\
\midrule
\textbf{Model} $\leftarrow$ \textit{PAWS EN Train Labeled by} \textcolor{azure(colorwheel)}{\textbf{Prompt}}
& \rotatebox[origin=c]{90}{\pawsx} 
& \rotatebox[origin=c]{90}{\stshard} 
& \rotatebox[origin=c]{90}{\mrpc} 
& \rotatebox[origin=c]{90}{\snli} 
& \rotatebox[origin=c]{90}{\anli} 
& \rotatebox[origin=c]{90}{\xnli} 
& \rotatebox[origin=c]{90}{\stseasy} 
& \rotatebox[origin=c]{90}{\sickeasy}
& \rotatebox[origin=c]{90}{\amr} 
& \rotatebox[origin=c]{90}{\simp} 
& Clfy! & Min! & Max! & \textbf{$\overline{Err}$}\\
\midrule %
\textbf{XLM-R} $\leftarrow$ \textit{Original Labels} (\textcolor{MutedGreen}{44$\%$} | \textcolor{MutedRed}{56$\%$})     
& \textbf{15.2} & \textbf{54.1} & 33.4 & \textbf{32.4} & \textbf{7.2} & \textbf{26.7} & 46.6 & 37.0 & 31.4 & 5.3 & \textbf{34.2} & \textbf{30.0} & 18.3 & \textbf{$^\star$27.5} \\ 
\textbf{XLM-R} $\leftarrow$ \textit{Llama3 \pa} (\textcolor{MutedGreen}{79$\%$} | \textcolor{MutedRed}{21$\%$})   
& 45.0 & 63.2 & 33.4 & 77.9 & 93.1 & 89.9 & 70.4 & 63.7 & 23.2 & \textbf{2.9} & 47.2 & 79.0 & \textbf{13.0} & 46.4 \\
\textbf{XLM-R} $\leftarrow$ \textit{Llama3 \pb} (\textcolor{MutedGreen}{59$\%$} | \textcolor{MutedRed}{41$\%$}) 
& 34.1 & 58.8 & \textbf{32.7} & 60.3 & 63.7 & 70.8 & 66.4 & 50.6 & 21.3 & 6.9 & 41.9 & 62.4 & 14.1 & 39.5 \\
\textbf{XLM-R} $\leftarrow$ \textit{Llama3 \pc} (\textcolor{MutedGreen}{56$\%$} | \textcolor{MutedRed}{44$\%$}) 
& 30.0 & 56.4 & 32.8 & 47.5 & 58.7 & 62.6 & 52.6 & 34.5 & 29.0 & 8.5 & 39.7 & 51.2 & 18.8 & 36.6 \\
\textbf{XLM-R} $\leftarrow$ \textit{Llama3} $\mathbf{\pa \land \pb \land \pc}$ & 28.4 & 54.6 & 34.0 & 33.7 & 49.0 & 55.7 & \textbf{40.6} & \textbf{19.0} & \textbf{38.1} & 10.7 & 39.0 & 39.6 & 24.4 & 34.3 \\
\bottomrule
\end{tabular}}
\caption{Ablation study B: XLM-R fine-tuned on PAWS Wiki training samples but instead using predicted Llama3 labels as targets (for description of prompts \pa, \pb, {\pc} see \ref{subsec:llm}). $X$ $\land$ $Y$: The positive label is assigned if and only if it is predicted by both prompt $X$ and $Y$. All results are error percentages: Lower numbers are better.  The classification method with the best overall performance is marked with an asterisk ($^\star$)}.
\label{tab:furtherresults}
\end{table*}

\section{Main Benchmark Results}

The main results on \ourbenchmark are shown in Table \ref{tab:mainresults}.

\paragraph{What model is best overall?} Interestingly, our most simple LLM setup (\pa) demonstrates the best overall result (21\% errors on average) on \ourbenchmark, outperforming the second best model that uses in-context examples from \pawsx (\paicl) by 3.2 percentage points (pp). While the smaller fine-tuned \textbf{XLM-R} model excels on \pawsx with an error rate of only 15.2\% it struggles to generalize across the other objectives and ranks second to last, showing an overall error of 27.5\%.

\paragraph{What model is best for specific objectives?} Findings on \ourbenchmark show that each model exhibits  unique strengths. When examining the averages for each objective (right side of Table \ref{tab:mainresults}), Llama with prompt \pcicl~(Min!, 0.8\% average error) outperforms others in the minimization objective, while Llama3 with prompt \pa~(Max!, 11.8\% average error) excels in the maximization objective.  In the classification objective, the trained \textbf{XLM-R} model achieves the best performance. However, it is important to note that this particular average is skewed due to the model's training on \pawsx. In the other two classification datasets, \textbf{XLM-R} is outperformed by Llama3 prompts, with up to a 14.8 pp error difference on \stshard (using prompt \pb) and a 9.8 pp difference on \mrpc (using prompt \pa).

\paragraph{Zero Shot or In Context Learning}
In the zero-shot experiments, the different paraphrase notions in the prompt lead to different levels of strictness and overall result, whereas in the ICL experiments the three prompts behave stricter. Adding in-context examples to \pa{} seems to calibrate the model to reduce false positives, but at the cost of a 19.8 pp error difference in the maximization average, and a lower overall result.

\section{Discussion}

\subsection{Can we improve the XLM-R training?} 

In the main results, we observed that fine-tuning a model on the \pawsx training set yields strong performance on the \pawsx test set. However, despite this success, the trained model ranks second to last overall on \ourbenchmark. A natural question then is: Why did the model fail to generalize?

\textbf{Is it the adversarial character?}  
We hypothesize that this may be due to  the adversarial ch-aracter of \pawsx. While its difficulty should ideally teach the model to handle even the most challenging cases, the artificial creation of the dataset may have introduced biases. These biases could have led the model to rely on spurious correlations -- similar to the "Clever Hans" effect -- where the model learns to exploit patterns that do not generalize well, potentially explaining its struggles with broader generalization \citep{niven-kao-2019-probing}.

\textbf{Ablation study A: Adding easy negatives to the training data.} To test this hypothesis, we introduce varying amounts of easy negative samples into the training data. To maintain the original label distribution, we generate these easy negatives by replacing one part of an adversarial negative pair with a random sentence from the training set.  The results are presented in Table \ref{tab:ablation_negatives_paws}. We observe that this strategy can indeed improve the generalization ability of the trained model, but only when the amount of easy negative samples are up to 50\%. Adding 25\% of easy negatives reduces the model's overall error from 27.5 to 26.3, with improvements on \sickeasy (-4.3 pp) and \xnli (-5.9 pp).

\textbf{Is it the labels?}
XLM-R models trained on \pawsx exhibit poor generalization on \ourbenchmark, which could be attributed to the target labels assigned to the adversarial sentence pairs.

\paragraph{Ablation Study B: Relabeling the Training Data with Llama3.}
To examine whether the model's generalization ability could be improved by utilizing labels generated by large language models (LLMs), we trained the model on the PAWS-Wiki training set using silver labels predicted by Llama3 through zero-shot prompting. Contrary to our hypothesis, the results in  Table~\ref{tab:furtherresults} show that models trained on these silver labels exhibit even poorer generalization performance, invalidating the assumption that relabeling enhances generalization. We eyeballed a handful of examples where all prompts agreed on the label, but disagreed with the original one. Some of these examples suggest annotation errors, such as the following: S1: \textit{In 1923, there were 568 wz.1902 guns in the Polish inventory} S2: \textit{In 1923, 568 Polish guns were in the inventory of wz.1902}. In this example, PAWS assigns a negative label, while all prompts assign a positive label. The consensus seems to suggest that the LLM has realized that `wz.1902' is a type of Polish gun, and all numbers do perfectly match. Though to be perfectly fair, an extremely strict interpretation could argue, e.g., that some of the guns in $S1$ might have been (e.g., accidentally) put in another inventory, hence supporting the assigned label, an insight that seems to ``post-hoc'' support our initial hypothesis in this experiment. But even when changing only those labels to positive where all prompts agreed on, the average error rises by about 6.8 pp.\ to 34.3\% (see Table \ref{tab:furtherresults}, last row).

\paragraph{Summary of ablation studies} 

What is the underlying cause of the model's lack of generalization? Based on our experiments, we cannot provide a definitive answer. The first experiment indicates that incorporating simple negative samples and reducing the adversarial nature of the training data may improve generalization. This adjustment also exposes the model to a broader spectrum of semantic similarity between sentence pairs, as opposed to the original dataset, which consisted primarily of difficult positives and negatives. On the other hand, the second experiment seems to indicate that the given PAWS labels are of mostly good quality. While apparently a few annotation mistakes can be corrected by LLM relabeling, any relabeling overall worsens a trained model's generalization.

We conclude that good performance on \pawsx does not always mean strong generalization in paraphrase detection overall. We have not yet identified a method to train an efficient model on PAWS that effectively generalizes to out-of-distribution scenarios. We speculate that over-fitting to complex adversarial data creation schemes may be at play, counteracting any potentially valuable learning signal (for now).

\subsection{Human Paraphrase Understanding Study}

\begin{figure}[t]
    \centering
    \includegraphics[width=\linewidth]{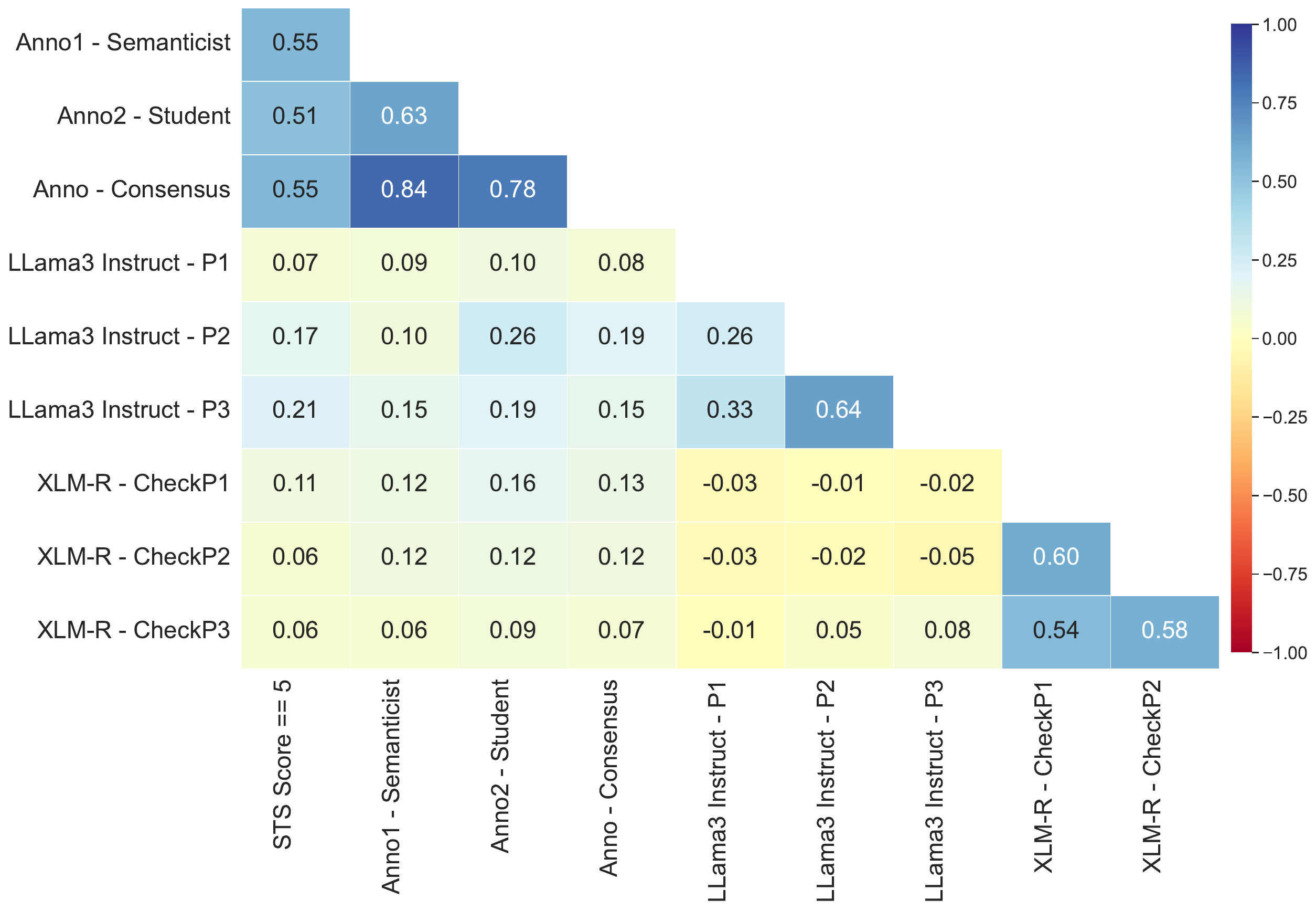}
    \caption{Cohens $\kappa$ between humans and systems when annotating the \stshard dataset that consists of highly similar (STS Score 4-5) sentences.}
    \label{fig:anno_heatmap}
\end{figure}

When creating \stshard, we obtained independent annotations from two annotators before agreeing on a final gold standard. One annotator was a researcher with a background in semantics, and the other was a student annotator. This setup allows for an interesting analysis of different agreement patterns: between the semanticist and the student, between the LLM and the trained model or one of the human annotators, and other comparisons.

We calculate pairwise Cohen's Kappa inter-annotator agreement scores \cite{Cohen:1960} and show the results in Figure \ref{fig:anno_heatmap}. We see that there is moderate to high agreement within humans and each model type. Then there is low agreement of Llama with human, and
basically zero agreement of XLM with both human and Llama. Interestingly, when considering the maximum STS label of 5 as indicative of a paraphrase, as intended by the task, we observe that human annotators exhibit only moderate agreement with this classification. Occasionally, both humans assigned a paraphrase to a pair from STS that has a score lower of 5.0, at other times, both humans agreed to assign a non-paraphrase label to an STS pair labeled with 5.0.


\subsection{Failure modes of LLM}

When analyzing LLM performance on \ourbenchmark using datasets containing only certified negatives or certified positives, we observed that while the error rate of the LLMs was low, it was never zero. 
Table  \ref{tab:examples} shows three pairs each from the repurposed NLI \xnli dataset (non-paraphrases only), as well as from the AMR-guidelines repurposed data (paraphrases only). In all these cases, all LLM prompts agree -- on the wrong decision. Interestingly, sometimes, the LLM seems to be tricked by seemingly simple active/passive variants and assigned a negative prediction, like \textit{X saddened Y} / \textit{Y was saddened by X} or a slight positional variation of an adverbial particle (\textit{look up X} / \textit{look X up}). At times, the model mistakenly labeled a contradiction as a paraphrase, as seen in the first pair where one sentence calls something a significant threat, while the other denies it was a threat.

Overall, we should note that LLMs generally show low error rates on the NLI-repurposed data, demonstrating robustness against hard negatives. However, the error rate is higher on hard positives in the AMR-repurposed data. In 92\% of the cases where one or more LLMs  make an error, at least one prompt still makes the correct decision. Nevertheless, our analysis indicates that LLMs can be misled by relatively simple linguistic variations.

\begin{table}[t]
    \centering
    \scalebox{0.69}{
    \begin{tabular}{l}
    \toprule 
\textbf{False Positive examples: LLMs predict paraphrase} \\
    \midrule
         \textit{The threat that was coming was not from sleeper cells.} \\
         \textit{Sleeper cells were the only threat of any significance.}\\
         \midrule
         \textit{The cover story reviews the latest research on how babies think.}\\
         \textit{The cover story talks about how infants make decisions.}\\
         \midrule
         \textit{(a) Change each d or t in the target to c.}\\
         \textit{After the conversion is finished the target should have exactly four c's.}\\
         \midrule
         \textbf{False Negative examples: LLMs predict non-paraphrase}\\ 
         \midrule
         \textit{The girl was saddened by the disaster.} \\
        \textit{The disaster saddened the girl.} \\
         \midrule
         \textit{Behavioral problems.}\\
        \textit{Problems behaving.} \\
         \midrule
         \textit{The boy looked up the answer.} \\
        \textit{The boy looked the answer up.} \\
\bottomrule
    \end{tabular}}
    \caption{Examples of LLM failures on NLI and AMR-repurposed data. In the first three pairs, all Llama3 prompts (\pa, \pb, and \pc) incorrectly classify the pairs as paraphrases, even though they are not. In contrast, in the last three pairs, the prompts fail to assign a paraphrase label, although all  examples are indeed paraphrases.}
    \label{tab:examples}
\end{table}

\section{Conclusion}

A key contribution of our work is the introduction of the \ourbenchmark benchmark, which includes sentence pairs from ten domains and covers three paraphrase detection challenges. We are releasing this benchmark to the research community, with the goal of facilitating further exploration of paraphrase phenomena, evaluating paraphrase detection models and zero-shot capabilities of LLMs, and pinpointing specific areas for improvement. 

Our experiments using \ourbenchmark revealed several interesting insights: 
\begin{enumerate}
\item 
 None of the tested LLM and classifier setups demonstrated strong performance  across the full spectrum of paraphrases  captured by the benchmark, highlighting its objective nature and the need for system development. 

\item Even modern LLMs as Llama3 failed to accurately and consistently detect  paraphrases  -- in some cases, the passivization of the  verb is all it takes to confuse all tested prompt variants.  

\item Improving training strategies for smaller, more efficient models is challenging, but not impossible. Specifically, carefully inserting a certain amount of easy negatives into a large training set of adversarial pairs helped to lower the average error. 

\item Among all tested LLM prompting strategies, each exhibited strengths and weaknesses depending  on the  particular challenge within  \ourbenchmark. However, the simplest prompt -- directly asking whether the sentence pairs are paraphrases -- performed best with an average error of 20.9\% across the benchmark.
\end{enumerate}

\section*{Limitations \& Future Work}

Our proposed benchmark includes text pairs with different levels of semantic and lexical similarity 
from diverse domains and covers many interesting paraphrasing phenomena. Does it cover all possible phenomena related to paraphrases and paraphrasing? Clearly not. For example, an obvious limitation is its coverage of phenomena in languages other than English. While
our dataset does include a total of 25,360 non-English samples, they are not spread across all ten parts of \ourbenchmark (e.g., the AMR-sourced dataset is only English). Thus, expanding \ourbenchmark to include more datasets in languages other than English, particularly low-resource languages, appears as a fruitful avenue for future work. We invite community collaboration to help extend \ourbenchmark.

While many resources were used to perform our evaluations, regarding our selection of LLMs, we limited ourselves to Llama3 Instruct 8B quantized to 4 bits. Similarly, the basic architecture for our experiments with trained models was limited to XLM-RoBERTa$_{base}$. We believe that the benchmark can be viewed as an extended reasoning evaluation of LLMs, useful for comparing capabilities of different sizes and architectures. Specifically, among all the many ablations and different setups that we ran for both model types, there was none that showed consistently good performance.

\section*{Acknowledgements}
This research is conducted under the project \textit{Impresso -- Media Monitoring of the Past II Beyond Borders: Connecting Historical Newspapers and Radio}. Impresso is a research project funded by the Swiss National Science Foundation (SNSF 213585) and the Luxembourg National Research Fund (17498891).

\bibliography{custom}

\appendix
\onecolumn
\newpage

\section{ICL Prompt}
\label{sec:prompts}

\begin{figure}[H]
\begin{tcolorbox}[title={In-Context-Learning (k=4) Paraphrase Detection Prompt},label={prompt-icl},colback=white]
Here are example sentence pairs that are \{paraphrase notion\} (Yes) and not \{paraphrase notion\} (No): \newline \newline
Sentence 1: \{RandPos1.sentence1\} \newline Sentence 2: \{RandPos1.sentence2\} \newline Yes \newline \newline
Sentence 1: \{RandNeg1.sentence1\} \newline Sentence 2: \{RandNeg1.sentence2\} \newline No \newline \newline
... \newline \newline
Are the following sentences \{paraphrase notion\}? \newline
\newline
Sentence 1: \{sentence1\} \newline
Sentence 2: \{sentence2\}
\newline
\newline
Answer with 'Yes' or 'No' 
\end{tcolorbox}
\caption{In-Context-Learning Prompt Template. For \pa, \pb, and \pc, the questions asked are ``paraphrases,'' ``semantically equivalent,'' and ``expressing the same content,'' respectively.}
\label{fig:prompt_icl}
\end{figure}

\section{Per Language Performance}
\label{sec:language}

\begin{table*}[h]
\centering
\small
\begin{tabular}{lrrrrrrr}
\toprule
\textbf{Model} & \textbf{en (E\%)} & \textbf{de (E\%)} & \textbf{fr (E\%)} & \textbf{es (E\%)} & \textbf{ja (E\%)} & \textbf{ko (E\%)} & \textbf{zh (E\%)} \\
\midrule
XLM-R$\leftarrow$ \textit{PAWS-EN Train} & 6.15 & 12.40 & 11.53 & 11.27 & 23.18 & 23.09 & 19.03 \\
Llama3 Instruct \pa & 34.75 & 43.90 & 43.30 & 43.60 & 49.70 & 50.20 & 47.25 \\
Llama3 Instruct \pb & 33.60 & 37.55 & 39.45 & 38.35 & 45.95 & 46.55 & 43.65 \\
Llama3 Instruct \pc & 29.65 & 35.00 & 35.85 & 34.90 & 45.00 & 45.15 & 40.85 \\
Llama3 Instruct \paicl & 35.50 & 37.65 & 38.40 & 39.10 & 40.25 & 43.50 & 38.25 \\
Llama3 Instruct \pbicl & 29.90 & 31.20 & 33.05 & 33.85 & 37.45 & 38.70 & 34.70 \\
Llama3 Instruct \pcicl & 27.95 & 31.15 & 32.20 & 32.85 & 36.55 & 38.50 & 33.30 \\
\bottomrule
\end{tabular}
\caption{Error rates of Table~\ref{tab:benchmark_overview} for the \pawsx test set per language.}
\label{tab:error_breakdown}
\end{table*}

\begin{table*}[h]
\centering
\small
\begin{tabular}{lrrrrr}
\toprule
\textbf{Model} & \textbf{en (E\%)} & \textbf{de (E\%)} & \textbf{fr (E\%)} & \textbf{es (E\%)} & \textbf{zh (E\%)} \\
\midrule
XLM-R$\leftarrow$ \textit{PAWS-EN Train} & 26.99 & 24.70 & 29.88 & 29.94 & 21.96 \\
Llama3 Instruct \pa & 15.18 & 12.71 & 13.16 & 11.87 & 8.74 \\
Llama3 Instruct \pb & 2.23 & 1.44 & 1.18 & 1.15 & 0.79 \\
Llama3 Instruct \pc & 2.40 & 1.24 & 1.14 & 1.20 & 0.63 \\
Llama3 Instruct \paicl & 4.82 & 3.10 & 2.60 & 1.98 & 1.44 \\
Llama3 Instruct \pbicl & 1.00 & 0.19 & 0.15 & 0.12 & 0.04 \\
Llama3 Instruct \pcicl & 0.78 & 0.13 & 0.15 & 0.13 & 0.06 \\
\bottomrule
\end{tabular}
\caption{Error rates of Table~\ref{tab:benchmark_overview} for the XNLI subset dataset for the five languages we use.}
\label{tab:error_comparison}
\end{table*}

\newpage

\section{Additional Datasets Statistics}
\label{sec:descriptive}

\begin{figure*}[h]
    \centering
    \fbox{\includegraphics[width=\textwidth]{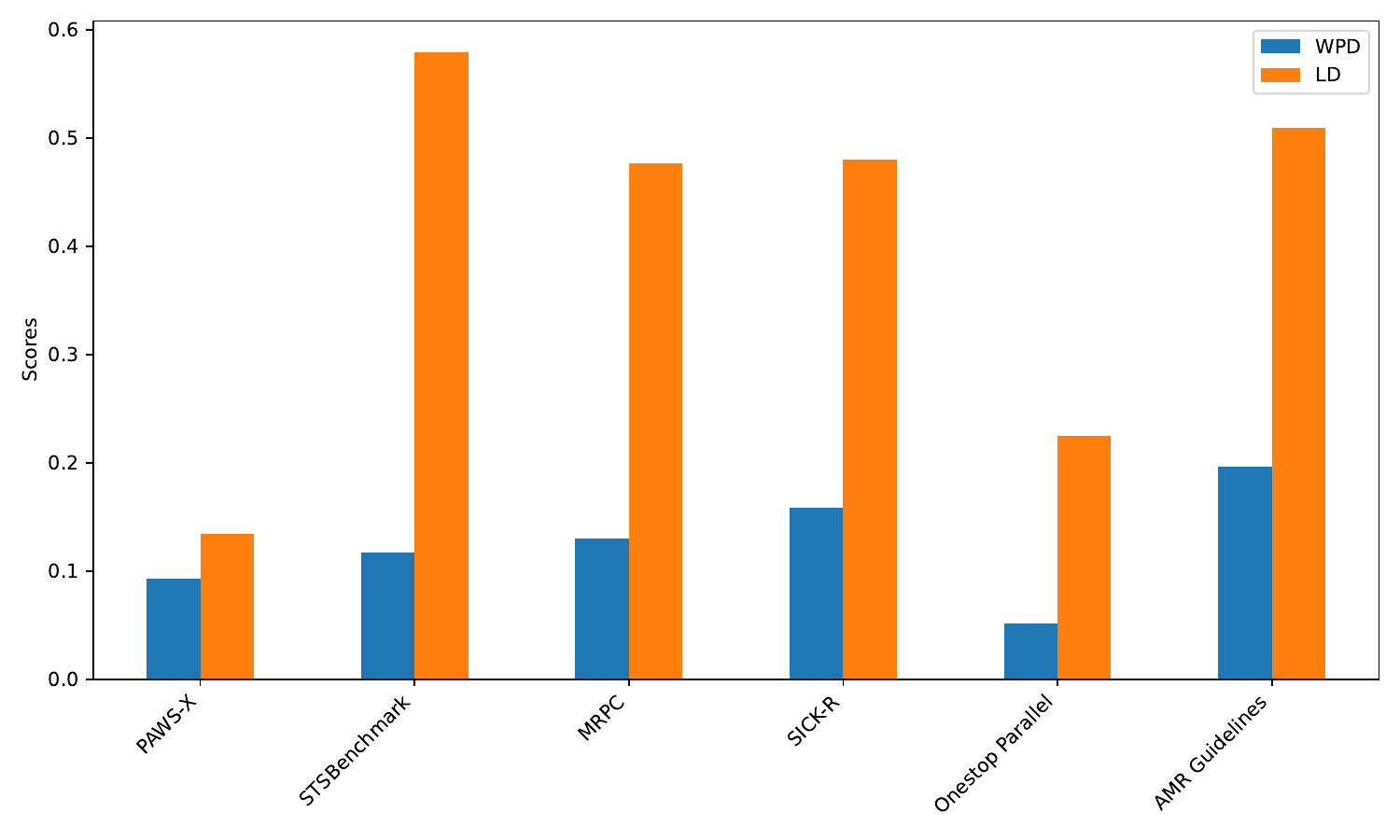}} 
    \caption{Average Word Position Deviation (WPD) and Lexical Diversity (LD) \citep{liu-soh-2022-towards} of the symmetric datasets of {\ourbenchmark}.}
    \label{fig:wpd_ld}
\end{figure*}

\begin{figure*}[b]
    \centering
    \fbox{\includegraphics[width=\textwidth]{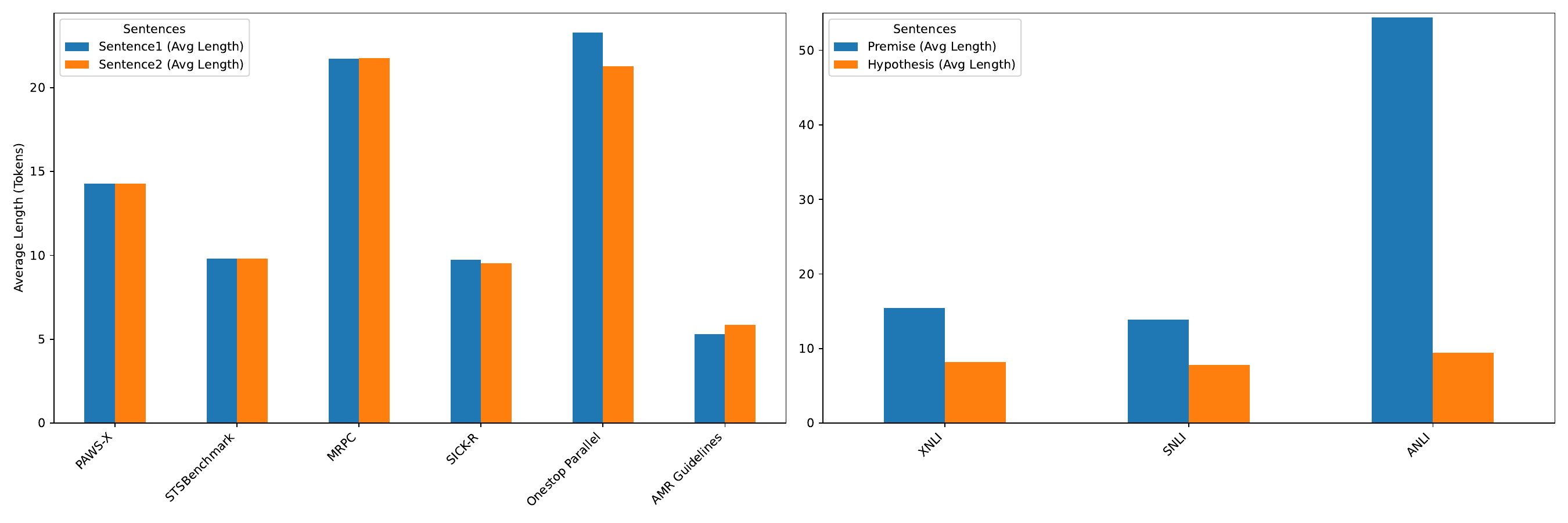}} 
    \caption{Average sentence lengths. Left plot shows symmetric datasets and right plot shows the asymmetric NLI datasets of {\ourbenchmark}}A
    \label{fig:sentence_length}
\end{figure*}

\end{document}